\documentclass{bmvc2k}

\usepackage{times}
\usepackage{epsfig}
\usepackage{graphicx}
\usepackage{amsmath}
\usepackage{amssymb}
\usepackage{multirow}
\usepackage{booktabs}
\usepackage{tabu}
\usepackage{mathtools}
\usepackage{url}
\usepackage{multirow,rotating}
\usepackage{multicol}
\usepackage{algorithm}
\usepackage{algpseudocode}
\usepackage[compatibility=false]{caption}
\usepackage{adjustbox}
\usepackage{dashbox}
\usepackage{xcolor,colortbl}

\DeclareMathOperator*{\argmax}{\arg\!\max}

\newcommand{\blt}{\ensuremath{\;\tiny{|}\;}}

\newcolumntype{a}{>{\columncolor{Gray}}c}
\newcolumntype{b}{>{\columncolor{white}}c}

\definecolor{Gray}{gray}{0.85}

\definecolor{red_}  {rgb}{ 1      0       0 }
\definecolor{blue_} {rgb}{ 0      0       1 }
\definecolor{pink_} {rgb}{ 1      0       0.83 }
\definecolor{gray_} {rgb}{ 0.6    0.6     0.6 }
\definecolor{cyan_} {rgb}{ 0.39   1       1 }
\definecolor{green_}{rgb}{ 0      1       0 }

\newcommand{\twopartdef}[4]
{
	\left\{
		\begin{array}{ll}
			#1 & \mbox{if } #2 \\
			#3 & \mbox{if } #4
		\end{array}
	\right.
}

\def\eg{\emph{e.g}\bmvaOneDot}

\def\etal{\emph{et al}\bmvaOneDot}

\newcommand{\RNum}[1]{\uppercase\expandafter{\romannumeral #1\relax}}

\title{Rule Of Thumb: Deep derotation for improved fingertip detection}

\addauthor{Aaron Wetzler}{www.cs.technion.ac.il/~twerd/}{1}
\addauthor{Ron Slossberg}{www.cs.technion.ac.il/~ronslos/}{1}
\addauthor{Ron Kimmel}{www.cs.technion.ac.il/~ron/}{1}

\addinstitution{
 Technion\\
 Israel Institute of Technology\\
 Haifa, Israel
}

\runninghead{Wetzler \etal}{Deep derotation for improved fingertip detection}

\begin{document}

\maketitle


\begin{abstract}
We investigate a novel global orientation regression approach for articulated 
 objects using a deep convolutional neural network. This is integrated with an in-plane image derotation scheme, DeROT, to tackle the problem of per-frame fingertip detection in depth images. The method reduces the complexity of learning in the space of articulated poses which is demonstrated by using two distinct state-of-the-art learning based hand pose estimation methods applied to fingertip detection. Significant classification improvements are shown over the baseline implementation. Our framework involves no tracking, kinematic constraints or explicit prior model of the articulated object in hand. To support our approach we also describe a new pipeline for high accuracy magnetic annotation and labeling of objects imaged by a depth camera. 
  
\end{abstract}


\section{Introduction} \label{sec:intro}

\begin{figure}[t]
\begin{center}
\includegraphics[width=0.98\linewidth]{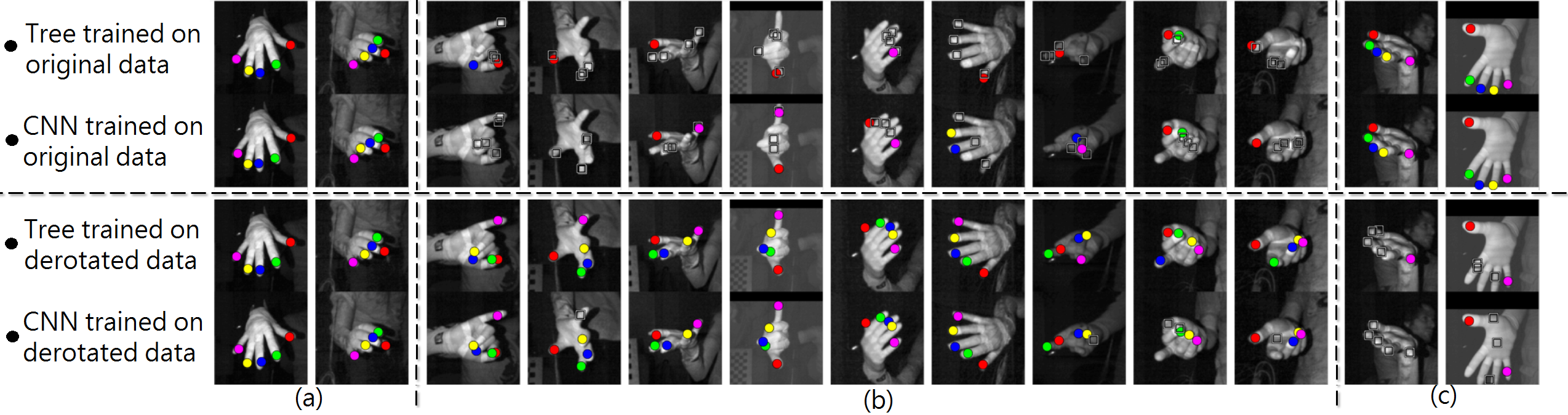}
\end{center}
   \caption{\scriptsize \color{blue} Examples from HandNet test set detections. The colors represent 
   fingertips that are correctly located and identified. The white boxes indicate false 
   detections with the error threshold chosen to be 1cm. The top two rows are trained and tested on 
   non-derotated data. The bottom two are trained and tested on derotated data and then rotated back 
   to the non-derotated space. The detections are overlaid on the IR image from the 
   camera which is not part of the classification process. a) Successful examples for all methods. 
   b) Representative challenging examples for which derotation enables better performance. 
   c) Failure cases where derotation fails to improve the results.}
   
\label{fig:fingertips}
\end{figure}

In this paper we propose a method for normalizing out 
the effects of rotation on highly articulated motion of deforming geometric surfaces such as 
hands observed by a depth camera. Changing the global rotation of an object directly increases the variation in appearance of the object parts. The work of \cite{KimHIBCOO12} 
physically removes this variability with a wristworn camera and samples only a single 3D point on 
each finger to perform full hand pose estimation. For markerless situations, removing variability 
through partial canonization can significantly reduce the space of possible images used for pose 
learning instead of trying to explicitly learn the variability through data augmentation. In 
\cite{LepetitLF05} the authors show that learning a derotated 2D patch instead of the original one 
around a feature point dramatically reduces the learning capacity required and improves the 
classification results while using fewer randomized trees. To develop our method we use fingertip 
detection as a challenging representative scenario with a propensity for self occlusion and high 
rotational variability relative to an imaging sensor. Many approaches in the literature use fingertip 
or hand part detection towards the goal of full 
hand pose (\eg \cite{KeskinKKA11},\cite{qian2014realtime},\cite{tompson14tog},\cite{Wang09}) 
however, they all approach the problem by trying to learn on datasets by augmenting rotational variability. 
Instead, we propose to remove  this hand space variability during both the training phase and run-time. To this end we propose to learn the rotation using a deep convolutional neural network (CNN) in a regression context based on a network similar to that of \cite{tompson14tog}. We show how this can be used to predict full three degrees of freedom (DOF) orientation information on a 
database of hand images captured by a depth sensor. We combine the 
predicted orientation with a novel in-plane derotation scheme. The "Rule of thumb" is derived from the following insight: there is almost always an in-plane rotation which 
can be applied to an image of the hand which forces the base of the thumb to be on the right side of 
the image. This implies that the ambiguity inherrent in rotationally variant features can be overcome 
by derotating the hand image to a canonical pose instead of augmenting a dataset with all variations 
of the rotational degrees of freedom as is commonly done. Figure \ref{fig:fingertips} shows examples of extensive
pose variation that can benefit from our approach \footnote{All graphs and images in this paper are best viewed in color.}. 

No currently available hand datasets (\eg \cite{ZhaoCX12},\cite{qian2014realtime},\cite{tompson14tog}) include accurate full 3 DOF ground truth hand orientations on a large database of real depth images. Using joint location data from 
NYUHands \cite{tompson14tog} it is possible to extract a global hand 
orientation per pose. However, we found that the size of this dataset and rotational variability are 
not optimal for learning to predict 3 DOF orientation. A significant contribution of this paper is 
therefore the creation of a new, large-scale database of fully annotated depth images with 212928 
unique hand poses captured by an Intel RealSense camera that we call HandNet\footnote{To advance research in the field this database and relevant code is available at \url{www.cs.technion.ac.il/~twerd/HandNet/}}. For the purpose of effectively 
annotating such a large dataset we describe a novel image annotation technique. To overcome the severe 
occlusion  inherrent in such a process we use DC magnetic trackers which are surprizingly sparsely 
used by the vision community considering their high accuracy, speed and robustness to occlusions.
Using our deep derotation method (DeROT) we show up to 20.5\% improvement in mean average precision (mAP) over our baseline 
results for two state-of-the-art approaches for fingertip detection in depth images, namely, a 
random decision tree \cite{KeskinKKA11} (RDT) and a deep convolutional neural network \cite{tompson14tog} (CNN). We 
also compare our results to a non-learning based method similar to PCA and show that it produces 
inferior results, further supporting the proposed use of DeROT.


\section{Building HandNet: Creation and annotation} \label{sec:database}

\begin{figure}[t]
\begin{center}
\includegraphics[width=0.95\linewidth]{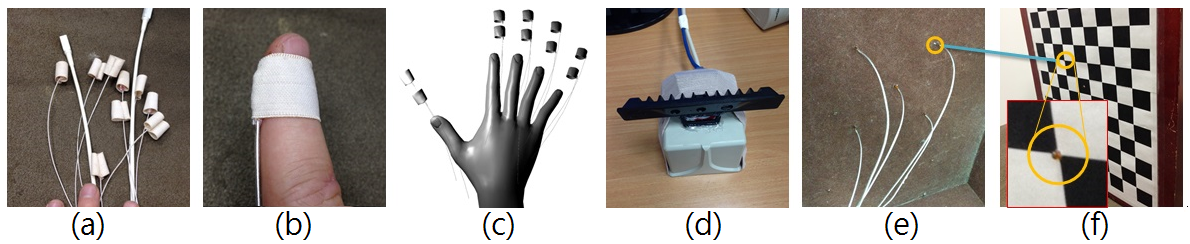}
\end{center}
   \caption{\scriptsize \color{blue} The data capture setup. a) 2mm magnetic sensors. The larger rectangular sensors are not used. b) A fingertip sensor inside the inner seam. c) Virtual model used for planning a multi-sensor setup. We only use 5 sensors. d) The RealSense camera rigidly fixed to the TrakStar transmitter. e) The back of the wooden calibration board where the glass sensor housings are firmly pushed through. f) The front of the calibration board where the glass sensor housings are visible on the corners as seen in the inset.}
\label{fig:dataprocess}
\end{figure}

Synthetic databases such as those created using \cite{libhand} have a severe disadvantage in that they cannot accurately account for natural hand motion, occlusions and noise characteristics of real depth cameras. 
The creation of a large hand pose database of real depth images with consistent annotations 
is therefore of great importance, but beyond the capability of human annotators. The NYUHands database \cite{tompson14tog} uses a full model of the hand and a three-camera setup to annotate hand joint locations. 
There are instances where fingers are obstructed and accurate orientation information 
is not reliable. Similarly the method of \cite{Wang09} uses inverse kinematics coupled with a colored glove which also has the disadvantage of not having explicitly measured orientation as well 
as fingertip locations which are obstructed from the depth camera. An alternative to model based 
systems are sparse marker systems such as those used by 
\cite{ZhaoCX12}, however, the excessive cost of a modern mocap setup 
such as Vicon as well as the occlusion problem make such an approach unattractive. 
In contrast, modern DC magnetic trackers like the TrakStar \cite{trakstar} are robust to metallic 
interference and obstruction by non-ferrous metals, and provide sub-millimeter and sub-degree accuracy for location and orientation relative to a fixed based 
station. Despite their almost non-existent use in modern computer vision literature, we have found them to be an excellent measurement and annotation tool. 

\textbf{Sensors.} To build and annotate our HandNet database we use a RealSense camera combined with $2mm$ TrakStar magnetic trackers. We affix the sensors to a user's hand and fingertips 
by using tight elastic loops with sensors in sewn seam pockets. This 
prevents lateral and medial movement along the finger. This can be seen in Figure \ref{fig:dataprocess}.  
The skin tight elastic loops have an additional significant benefit over gloves in that the depth 
profile and hand movements are not affected by the attached sensors and thus do not 
pollute the data.

\textbf{Callibration.} Camera callibration with known correspondences is a well studied problem \cite{Zhang96}.
However, in our case we need to callibrate between a camera and a sensor frame.
We do this by positioning the magnetic sensors on the corners of a checkerboard 
pattern thereby creating physical correspondence between the detected corner locations and 
the actual sensors. This setup can be seen in Figure \ref{fig:dataprocess}. We use the extracted 
2D locations of the corner points on the callibration board \cite{bouguet2004camera} together with the sampled sensor 3D locations to perform  EPnP \cite{Epnp09} to determine the extrinsic configuration between the devices.

\begin{figure}[t]
\begin{center}
\includegraphics[width=0.9\linewidth]{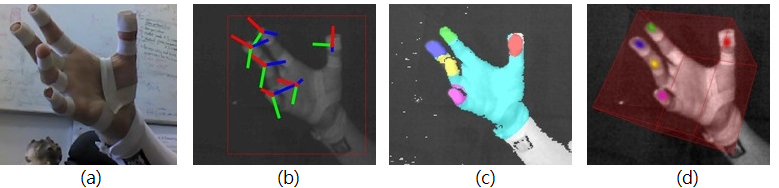}
\end{center}
   \caption{\scriptsize \color{blue} The available data annotations after calibration. a) Color image. Illustrates a full hand setup for this work. The color is not used. b) The RGB axes indicate the measured location and orientation of each fingertip and the back of the palm. c) IR image(not used) overlaid with the labels generated from the raycasting described in Section \ref{sec:database}. d) IR image overlaid with the generated heatmaps per fingertip and the global orientation of the hand represented as an oriented bounding box (not used).  }
\label{fig:annotation}
\end{figure}

\textbf{Annotation.} We model each sensor as a 3D oriented ellipsoid. We then raycast the ellipsoid into the camera frame and set the label to be the identity of the ellipsoid closest to the camera for every pixel. We also create a heatmap $h_i$ for each fingertip $i$ using the same technique but setting the value per pixel to be gaussian over the distance to the projected sensor location. An example of both types of annotation can be seen in Figure \ref{fig:annotation}.

\textbf{Recording the database.}
The database is created from $10$ participants (half male, half female, different hand sizes) who perform random hand motions with extensive pose variation while wearing the magnetic sensors. The RealSense camera operates at $~58$fps producing $640\times 480$ depth maps which we reduce to $320\times 240$. The TrakStar samples measurements at a rate of $720$Hz. In total we recorded $256987$ images. A portion of these images were removed because of low quality. The final dataset is $212928$ frontal images including full annotation of the position and orientation of each fingertip and the back of the palm. After recording each participant we used a software utility to add offsets to the rotation and location of each sensor to adjust for greater consistency in positioning across subjects.
\section{Fingertip detection} \label{sec:fingertip}
Although there are many non-learning based hand pose methods that can produce fingertip locations (\eg \cite{schmidt2014dart,oikonomidis2011markerless,MelaxKO13,BallanTGGP12}), they use kinematic and frame to frame constraints coupled with hand modelling. In contrast, here we specifically focus on per frame fingertip detection in depth images without either tracking or kinematic modelling. For our pipeline we first segment the target hand from the depth image using a fast depth based flood-fill method seeded either from the previous frame for real-time use and testing or from the ground truth hand location for building the database. Using the center of mass (CoM) of the segmented hand and its average depth value we define a depth dependent bounding box of size $w=\frac{50000}{z}$ for a RealSense camera (HandNet) and $w=\frac{70000}{z}$ for a Kinect camera (NYUHands) where $z$ is the depth of the CoM of the segmented hand. We derotate the image about the CoM using an angle of rotation according to the in-plane angle produced by DeROT described in Section \ref{sec:derotation}. This comes from the predicted full 3D orientation at run-time or from the ground truth sensor orientation for database construction or testing. We then crop the image using the bounding box. We now describe our modifications of the two different, learning-based fingertip detectors that we use in this work.
\begin{figure}[t]
\begin{center}
\includegraphics[width=0.8\linewidth]{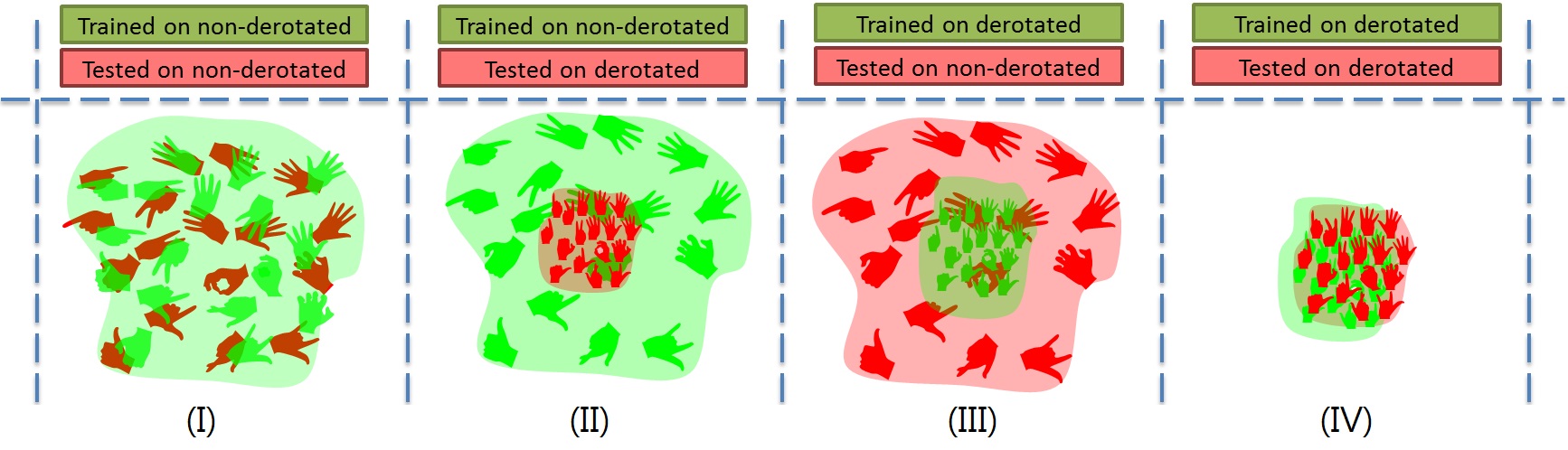}
\end{center}
   \caption{\scriptsize \color{blue} Understanding derotation: We represent the space of poses by a non-uniform 2D region with representative hand poses. Red and green represent the pose-space covered by training images and testing images respectively. Each \RNum{1},\RNum{2},\RNum{3},\RNum{4} indicates one of the 4 possible combinations of training and testing for a machine learning method where the database remains fixed in size. The larger region indicates greater pose variability while the smaller represents less. Intuitively, by training on a space with low variance and testing in this same space (type \RNum{4}) we expect to see an improvement over the opposite (type \RNum{1}). Section \ref{sec:evaluation} supports this intuition. }
\label{fig:ronnyhands}
\end{figure}
\subsection{Random decision tree} \label{sec:RDT}
We follow the method of Keskin \etal \cite{KeskinKKA11} where a random decision tree (RDT) ensemble learns hand part labels for every pixel in a depth image of a hand. We refer the reader to the supplementary material of our paper as well as \cite{KeskinKKA11, Shotton11} for specific details of this approach. However, here we propose a number of key differences which we found specifically helpful for fingertip detection and run-time efficiency.  We use the same random binary depth attributes per pixel but spatially distribute them according to an exponential sampling pattern similar to that of BRISK \cite{leutenegger2011brisk}. In addition to this, we use only a single RDT which contrasts with the common use of multiple trees in an ensemble. After training our single RDT the class distributions stored at each leaf can be used for inference because they represent the empirical estimate of the posterior probability $p\left(c | x \right)$ of hand part label $c$ given the image evidence $x$. Inferring the most likely fingertip identity label is therefore simply performed pixel-wise by finding the $c^*$ which maximizes $p\left(c | x \right)$ per pixel. However, label inference performed this way results in noisy labels as neighboring classifications do not influence one another. Without adding more trees we propose a simple but highly effective spatial regularization: for each fingertip $i$ we treat the posterior $p\left(c=i | x \right)$ for all pixels as an image and convolve it with a discrete 2D gaussian smoothing kernel $g_\sigma$ with blur radius $\sigma$.  This has the effect of correlating the posterior label distributions of nearby pixels. Therefore every pixel $q$ is labeled by fingertip identity (including palm and wrist labels)  according to
\begin{equation}\label{eq:treeEq}
c^*\left(q;x\right) = \mathop {\arg \max }\limits_{i \in \left\{0..6 \right\}} \left(g_{\sigma}* p_{c=i|x} \right) \left(q\right).
\end{equation}
Finally, we found that the close proximity of fingers compromises standard mean-shift \cite{meanshift} clustering. Instead we detect the largest label blobs in the label image from Equation \ref{eq:treeEq} above a certain area threshold. The 2D fingertip locations are then assigned to the blob centers and, if necessary, the average depth value for each blob can be used to generate the 3D camera-space coordinates. 


\textbf{Training the RDT.}
Training optimal decision trees is known to be NP-complete \cite{HyafilR76} and therefore trees are built from the root down using breadth-first greedy optimization over tree node impurity. We use the Gini impurity measure which is slightly cheaper to compute than the more typical entropy measure. To build our database for training an RDT we extracted $80\%$ of the fingertip pixels in our training datasets and $50\%$ of the non-fingertip hand pixels. For HandNet this results in a training dataset of $500$ million sample pixels totaling ~$600$GB of data for $1200$ attributes. Our tree-builder trains an unpruned randomized tree on $4\times$ GTX 580 GPUs and an Intel I7  processor with $48$GB of RAM in $16$ hours for a tree depth of $21$ with $18000$ query tests per node. We are not aware of another single-workstation tree-builder capable of handling this quantity of data. The very large number of examples helps to prevent overfitting demonstrated by single RDTs.


\subsection{Convolutional neural network} \label{sec:CNN}
For our second evaluated method we build a CNN architecture based on Tompson \etal \cite{tompson14tog} to predict the location of the five fingertips by using the maximum location in a set of heat maps which implicitly represent fingertip locations. We refer the reader to that work for specific details and to our supplementary material for the explicit architecture of our implementation. This multi-layer deep approach is critical for an input space as 
complicated as the set of images of an articulated object and we found that the deeper convolutional layers extract feature responses on a higher semantic level such as oriented fingertips. Using the heatmap based error objective helps to spatially regularize the network during training. For input to the CNN we set $D_1$ to be the cropped depth resized to $96\times 96$ pixels. We then downsample it by a factor of two twice to produce $D_2$ and $D_3$. We use a subtractive form of local contrast normalization (LCN) \cite{tompson14tog,jarret} so that $D_i \leftarrow D_i - g_{\sigma}*D_i$ using a gaussian smoothing kernel with $\sigma=5$ pixels. The triplet $\left(D_1,D_2,D_3\right)$ is then input to the network. The trained network outputs a heatmap $h_i$ per fingertip $i$ for new data. Our method differs for fingertip detection in that we augment the output by a non-fingertip heatmap that is strong wherever a fingertip is not likely to be present. Also, instead of fitting a gaussian model to the strongest mode in the low resolution heatmaps, we instead upsample each $18\times 18$ fingertip heatmap $h_i$ to a fixed size of $128\times 128$ with a smoothing bi-linear interpolator. Similar to Section \ref{sec:RDT} every pixel $q$ is labeled with fingertip identity (including a non-fingertip class)
\begin{equation}\label{eq:cnnEq}
c^*\left(q\right) = \mathop {\arg \max }\limits_{i \in \left\{0 .. 5 \right\}} h_i\left(q\right).
\end{equation}
As in Section \ref{sec:RDT} the fingertip locations are given by the location of the largest label blob.

\textbf{Training the CNN.} 
Both the orientation regression CNN of the next Section as well as the described fingertip CNN are trained using Caffe \cite{jia2014caffe} on an NVidia GTX 980 with an i7 processor and $16$GB of onboard RAM. We train both with a Euclidean loss and a batch size of $100$ for $100000$ iterations with stochastic gradient descent. We start with a learning rate of 0.01 and reduce it by a factor of 0.2 after every ten thousand iterations. We found that repeated fine-tuning was necessary to help network convergence.


\section{Derotation}\label{sec:derotation}

\subsection{Orientation regression}\label{sec:regression}

\begin{figure}[t]
\begin{center}
\includegraphics[width=0.9\linewidth]{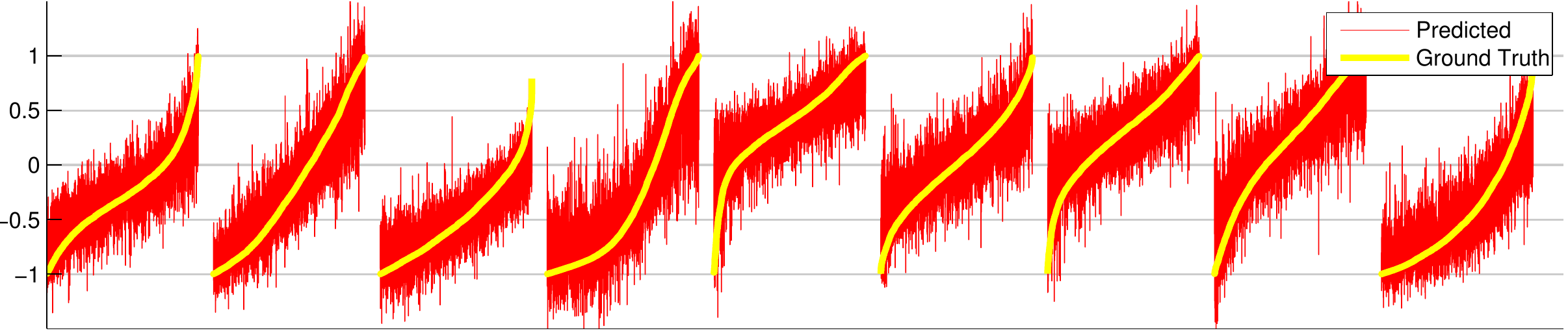}
\end{center}
   \caption{\scriptsize \color{blue} This graph shows the predicted value of all 9 coefficients of the hand orientation matrix in red relative to the ground truth in yellow. For clarity we order each ground truth coefficient monotonically and apply this reordering to the predicted results. The mean squared error for all the coefficients on the HandNet test set before and after SVD is 0.0271 and 0.0234 respectively.}
\label{fig:rotationgraph}
\end{figure}

We adapt the deep convolutional architecture from Section \ref{sec:fingertip} to predict full $3$ DOF hand orientation. Instead of a heatmap, we directly predict the $9$ coefficients of the rotation matrix. There are only $3$ degrees of freedom in a regular rotation but by using $9$ parameters and a large database we are effectively regularizing our over-parameterized output. The representation of a rotation matrix in this way is unique in $SO\left(3\right)$ unlike quaternions and Euler angles which we found to be noisy and unreliable. This noise was most visible when trying to predict a single representative angle. For training we use Euclidian loss and do not enforce orthonormality. However, the output of this CNN is directly projected onto the closest unitary matrix using the SVD decomposition $R = USV^T$. $\hat{R} = UV^T$ then provides a least squares optimal projection into $SO\left(3\right)$, if we additionally enforce $\det\left(\hat{R}\right)=1$. Figure \ref{fig:rotationgraph} shows the result of predicting the $9$ ground truth coefficients for HandNet and the full network architecture can be seen in the supplementary material. 


\subsection{DeROT: Designing a derotation method}\label{sec:derot}

\begin{algorithm}[h]
\scriptsize
\begin{multicols}{2}
\begin{algorithmic}[1]
\Procedure{Derotate}{$R$}   
	\State $r_{align} \leftarrow \argmax_{r_i \in \{r_1,r_2,r_3\}} \|\left(0,0,1\right)\cdot r_i\|$
	\If {$r_{align}$ = $r_2$ (thumb aligned axis) }
	   \State $\alpha \leftarrow atan2(r_{3x},r_{3y}) + 90 + \tiny \twopartdef {180} {r_{2z} \leq 0} {0} {r_{2z} > 0}$
	\Else
	   \State $\alpha \leftarrow atan2(r_{2x},r_{2y}) + 90$	   
	\EndIf
	\State Return $\alpha$
\EndProcedure
\end{algorithmic}


\caption{Derotation procedure}\label{alg:treebuilder}
\label{alg:derot}
\end{multicols}
\vspace*{-0.3cm}
\end{algorithm}

\begin{figure}[h]
\begin{center}
\includegraphics[width=0.95\linewidth]{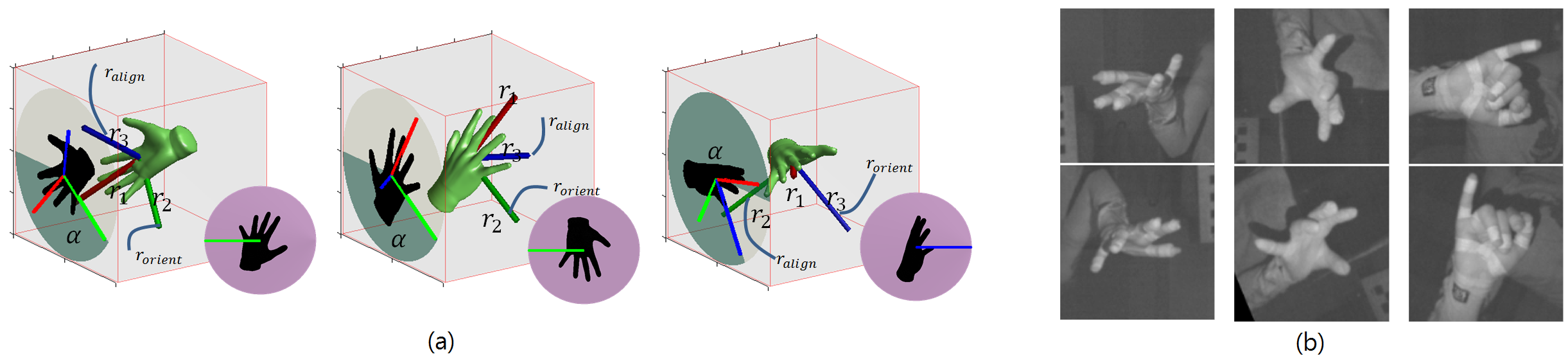}
\end{center}
   \caption{\scriptsize \color{blue} Synthetic and real examples of DeROT. a) The depth projection of the virtual hand before applying DeROT can be seen on the left wall of the cube representing the camera plane. The axis marked $r_{orient}$ is projected onto the camera plane and used in DeROT to define the angle $\alpha$. The purple circle contains the resulting image of the hand after applying derotation by angle $\alpha$. b) The top row of images are un-derotated. The bottom row have been derotated by $\alpha$ obtained by DeROT. Note that the thumb is consistently on the right of the image.}
\label{fig:derot}
\end{figure}


We take advantage of the orientation prediction $\hat{R}=\left[r_1 r_2 r_3\right]$ to compute an angle $\alpha$ which we will use for rotating the camera image about its center. The aim of this is to reduce pose variance by heuristically forcing the thumb to be on the right side of the image. We could use a predefined axis and set the angle $\alpha$ with which to rotate the image to be that between the projection of this axis and the upwards image direction. Unfortunately, when this axis mostly points to or away from the camera the projection onto the screen will be small and noisy. As a simple heuristic we detect if this is the case and if so choose an alternative axis. Specifically we first determine the predicted axis $r_{align}$ most aligned with the camera z axis as $r_{align}=\argmax_{r_i \in \{r_1,r_2,r_3\}} \|\left(0,0,1\right)\cdot r_i\|$. If $r_{align}$ is either the palm pointing direction or the direction of the extended fingers then we can be sure that the thumb direction $r_2$ will be non-noisy for this case and set $r_{orient}=r_2$. If the test yields instead that $r_{align}=r_2$ (i.e. thumb direction is mostly pointing towards or away from the camera) then we instead set $r_{orient}=r_3$ which is the palm vector. This procedure is summarized in Algorithm \ref{alg:derot}. Synthetic and real examples can be seen in Figure \ref{fig:derot}. This choice is arbitrary and can be adapted for objects other than the hand. We thus define DeROT to be the combination of using the CNN from Section \ref{sec:regression} to predict $\hat{R}$ together with this derotation heuristic. 


\subsection{Derotation with PCA and Procrustes} \label{sec:pca}
Instead of using DeROT, an alternative approach is to extract the principal axes of the hand silhouette using PCA and taking the rotation angle of the largest axis to the vertical image axis. We have found that a similar but more stable option is to determine an enclosing ellipse using a Procrustes like algorithm on the convex hull of the points $\cal V$ of the hand segmentation. The minimum area enclosing ellipse can be found efficiently over the points $x_i\in\text{convhull}\left({\cal V}\right)$ by minimizing $-\log \left( {\det \left( A \right)} \right) \text{,  s.t  } {\left( {{x_i} - {\overline x_i}} \right)^T}A\left( {{x_i} - {\overline x_i}} \right)\leq 1$ for $A,{\overline x_i}$ defining the ellipse. We solve this using Khachiyan's algorithm \cite{AspvallS80}. However, as shown in Section \ref{sec:evaluation} even with this added stability the method reduces performance rather than improving it. 
 

\section{Experiments} \label{sec:experiments}


\subsection{Evaluation protocol and data} \label{sec:evaluation}

\textbf{Experiments.} We perform our experiments using our HandNet database and the publicly available database NYUHands \cite{tompson14tog}. All experiments are performed separetly on the two databases. Our baseline results come from (\RNum{1}) training on non-derotated data and testing on non-derotated data. We compare this to (\RNum{2}) training on non-derotated data while testing with derotated data, (\RNum{3}) training on derotated data while testing with non-derotated data, (\RNum{4}) training on derotated data while testing with derotated data. 

\textbf{Non-derotated data.} For HandNet training we randomly select 202928 images and use the remaining 10000 images for testing. For NYUHands we use all 3 camera views (72757 images per view) for training and the frontal view for testing (8252 images). We slightly dilute the training and testing sets according to our hand segmentation pipeline which results in 184100 training images and 7241 testing images. 

For experiment types (\RNum{2}) and (\RNum{4}) we use this data to train two CNN orientation regression networks; one for each dataset. We use the same data for training the RDT and CNN fingertip detectors for experiment types (\RNum{1}) and (\RNum{2}). However, for testing the fingertip detectors in experiments (\RNum{1}) and (\RNum{3}), we rotate each testing image by uniformly random in-plane rotational offsets between -$90$ and $90$ degrees. This further guarantees that the testing data is different from the training data. 

\textbf{Derotated data.} Experiment types (\RNum{3}) and (\RNum{4}) use training data which is first derotated by an Oracle which we define to be DeROT that uses the ground truth $R_{gt}$ obtained from the magnetic sensors. With experiment types (\RNum{2}) and (\RNum{4}) we first apply the same uniform random image rotation to the test images exactly as for experiment types (\RNum{1}) and (\RNum{2}). We then apply one of the following: (a) Procrustes derotation, (b) DeROT using $\hat{R}$ predicted by the CNN regression network, (c) Oracle derotation with $R_{gt}$. 

\textbf{Mean precision and mean average precision.} We compute precision and recall according to the protocol of \cite{voc11}. We set prediction confidence as the value at the location of the fingertip detection in the $128\times 128$ channel heatmap for each fingertip. The mean precision (mP) represents the mean precision over all fingertips at a recall rate of $100\%$. Mean average precision (mAP) measures the mean of all the areas under the precision-recall curves for each fingertip and takes into account the behaviour over all confidence values. 

\textbf{Error threshold.} The error of a prediction is the distance to the ground truth location. If a fingertip is more than 6 pixels from the ground truth position it is considered a false positive. The threshold of 6 pixels roughly translates into a distance of $1cm$ for both HandNet and NYUHands in an image patch of size $128\times 128$ cropped according to Section \ref{sec:fingertip}. $1cm$ is a natural threshold to choose as the distance between adjacent fingertips is over $1.6cm$ on average \cite{dandekar2003}. 


\begin{table}[htbp]
\scriptsize
  \centering  %
  \
    \begin{tabular}{rrrrr}     
    \toprule

    \multicolumn{1}{c}{Test set derotation method} & \multicolumn{1}{c}{None} & \multicolumn{1}{c}{(a) Procrustes} & \multicolumn{1}{c}{(b) DeROT} & \multicolumn{1}{c}{(c) Oracle} \\
    \multicolumn{1}{c}{} & \multicolumn{1}{c}{\tiny mP \blt mAP} & \multicolumn{1}{c}{\tiny mP \blt mAP} & \multicolumn{1}{c}{\tiny mP \blt mAP} & \multicolumn{1}{c}{\tiny mP \blt mAP} \\
    
    \midrule
    
    \multicolumn{5}{c} {\normalsize HandNet} \\
    
    \midrule

    \multicolumn{1}{l}{ RDT trained on non-derotated data} &  
    \cellcolor{red_!50} {\underline{0.51} \blt \underline{0.79}  } & 
    \cellcolor{pink_!30}  {\underline{0.49} \blt \underline{0.77}} & 
    \cellcolor{pink_!30}{\textbf{0.55 \blt 0.85}} & 
    \cellcolor{pink_!30} {\textbf{0.60 \blt 0.87} } \\

    \multicolumn{1}{l}{ RDT trained on derotated data} & 
    \cellcolor{blue_!40}{0.32 \blt 0.60 } &  
    \cellcolor{cyan_!40}{} & 
    \cellcolor{cyan_!40}{\textbf{\underline{0.63}  \blt \underline{0.88}}} & 
    \cellcolor{cyan_!40}{\textbf{\underline{0.75}  \blt \underline{0.95}} } \\
    
    \midrule 

    \multicolumn{1}{l}{ CNN  trained on non-derotated data} &  
    \cellcolor{red_!50}{\underline{0.44}  \blt \underline{0.73}} & 
    \cellcolor{pink_!30}{\underline{0.42}  \blt \underline{0.73}} &     
    \cellcolor{pink_!30}{\textbf{0.46  \blt 0.77}} & 
    \cellcolor{pink_!30}{\textbf{0.50  \blt 0.79}} \\
    
    \multicolumn{1}{l}{ CNN  trained on derotated data} & 
    \cellcolor{blue_!40}{0.30  \blt 0.59} &  
    \cellcolor{cyan_!40}{} & 
    \cellcolor{cyan_!40}{\textbf{\underline{0.61}  \blt \underline{0.88}}} & 
    \cellcolor{cyan_!40}{\textbf{\underline{0.74}  \blt \underline{0.95}}} \\
    
    \midrule
    
    \multicolumn{5}{c} {\normalsize NYUHands} \\
    
    \midrule

    \multicolumn{1}{l}{ RDT trained on non-derotated data} &  
    \cellcolor{red_!50} {\underline{0.51}  \blt \underline{0.75}} & 
    \cellcolor{pink_!30}  {\underline{0.47}  \blt \underline{0.73}} & 
    \cellcolor{pink_!30}{\textbf{0.58  \blt 0.84}} & 
    \cellcolor{pink_!30} {\textbf{0.61  \blt 0.86}} \\

    \multicolumn{1}{l}{ RDT trained on derotated data} & 
    \cellcolor{blue_!40}{0.35 \blt  0.58 } &  
    \cellcolor{cyan_!40}{} & 
    \cellcolor{cyan_!40}{\textbf{\underline{0.63}  \blt \underline{0.88}}} & 
    \cellcolor{cyan_!40}{\textbf{\underline{0.68}  \blt \underline{0.89}}} \\
    
    \midrule 

    \multicolumn{1}{l}{ CNN  trained on non-derotated data} &  
    \cellcolor{red_!50}{\underline{0.38}  \blt \underline{0.70}} & 
    \cellcolor{pink_!30}{\underline{0.36}  \blt \underline{0.69}} &     
    \cellcolor{pink_!30}{\textbf{0.46 \blt  \underline{0.80}}} & 
    \cellcolor{pink_!30}{\textbf{0.48 \blt  \underline{0.81}}} \\
    
    \multicolumn{1}{l}{ CNN  trained on derotated data} & 
    \cellcolor{blue_!40}{0.23 \blt  0.42} &  
    \cellcolor{cyan_!40}{} & 
    \cellcolor{cyan_!40}{\textbf{\underline{0.49}  \blt 0.72}} & 
    \cellcolor{cyan_!40}{\textbf{\underline{0.53}  \blt 0.73}} \\    
    
    \bottomrule
    
    \end{tabular}%
    \
     \caption{\scriptsize \color{blue}{Results of our experimental evaluation for all experiment types described in Section \ref{sec:evaluation}. The experiment types \RNum{1},\RNum{2},\RNum{3},\RNum{4} are highlighted in red, pink, blue and cyan respectively. A result in bold indicates that it outperforms the baseline (\RNum{1}) shown in red. For each row pair (derotated training data vs non-derotated training), the underlined result is the better of the two. Procrustes consistently reduces the quality of fingertip detection. Conversely, DeROT outperforms the baseline \emph{for every experiment}. For all but one experiment, this improved performance is significantly enhanced by training on derotated data instead of original data. See Section \ref{sec:discussion}. The results from the Oracle serve as an upper bound achievable by derotation. }}
  \label{tab:HandNet}%
\end{table}%


\begin{figure}[ht]
\begin{tabular}{cc}
\centering     
\subfigure[RDT]{\includegraphics[width=0.45\linewidth]{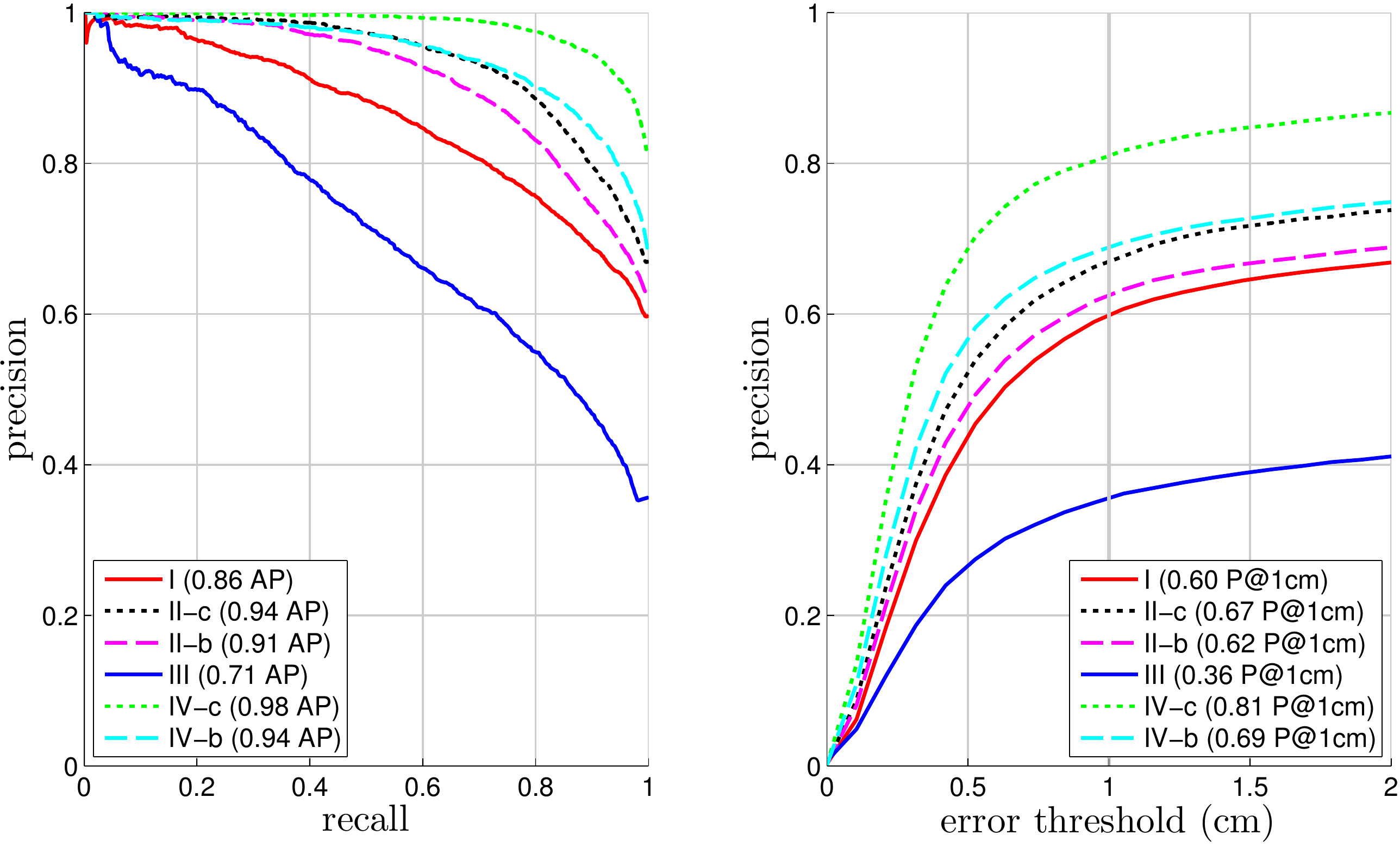}} &
\subfigure[CNN]{\includegraphics[width=0.45\linewidth]{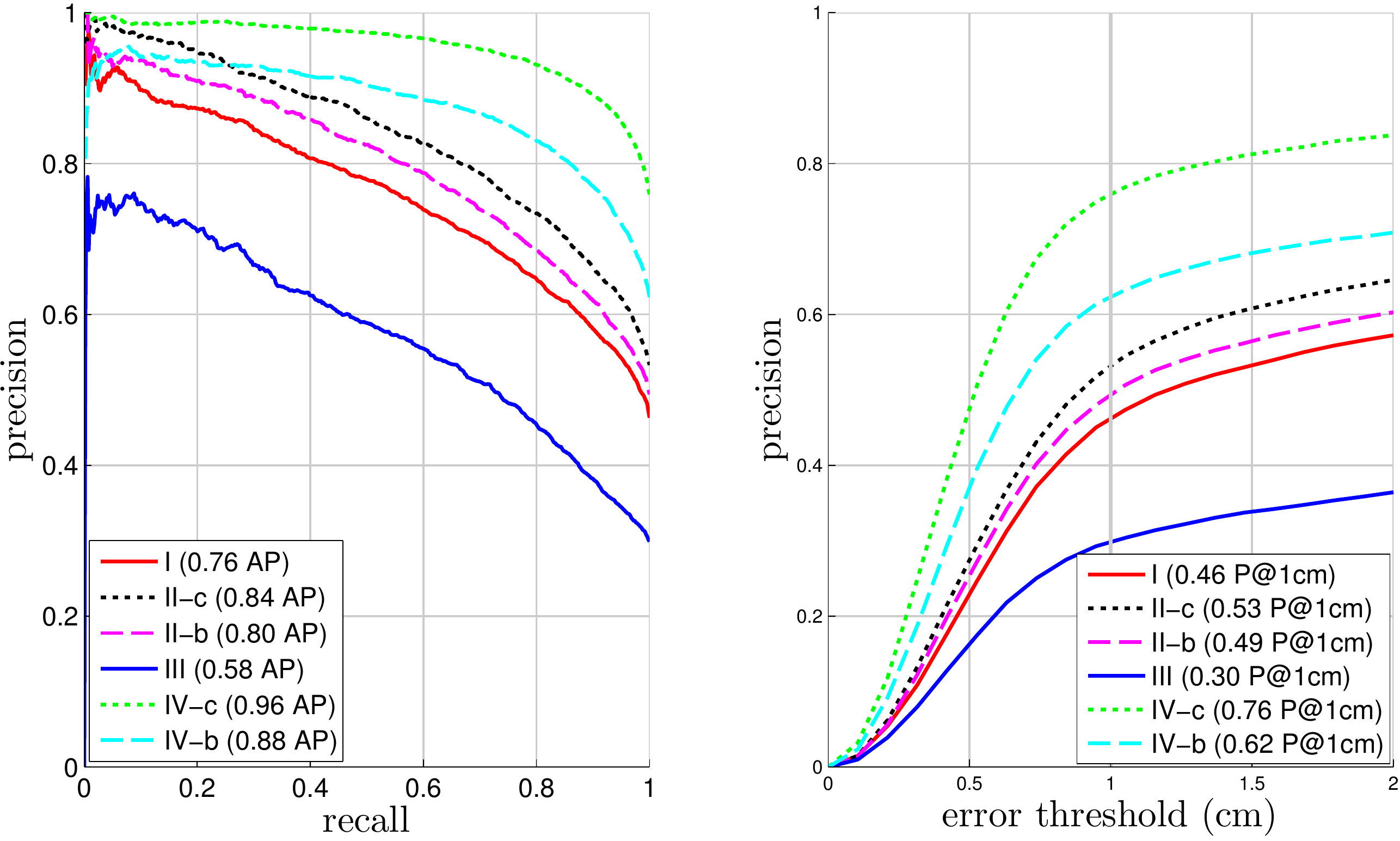}} \\
\end{tabular}
\caption{\scriptsize \color{blue} These graphs show typical precision to recall and precision to error threshold for thumb detection (using RDT and CNN on the HandNet test set. Each line indicates an experiment which is labeled in the legend using the experiment types from Section \ref{sec:experiments} and the derotation types Procrustes(\emph{a}), DeROT(\emph{b}), Oracle(\emph{c}). The baseline is in red. Training on derotated data and then applying DeROT or Oracle is in cyan or green respectively. Training on non-derotated data and then applying DeROT or Oracle is in magenta or black respectively. The average precision (AP) and precision at 1cm error (P@1cm) are shown for each thumb experiment.}
\label{tab:graphs}
\end{figure}


\subsection{Discussion}\label{sec:discussion}
The results of the experiments can be seen in Table \ref{tab:HandNet}. In Figure \ref{tab:graphs} we display a precision-recall curve and error threshold graph for the thumb on the HandNet test-set for all experiment types which is representative of the behavior of all fingertips. The results show that the use of DeROT improves over the baseline results \emph{for all measurements} for both RDT and CNN for experiments on both datasets. On HandNet, when training an RDT and CNN on ground truth derotated data, we see that test-time use of DeROT yields improvement in mAP of 11.3\% and 20.5\% over the respective baselines. For NYUHands, DeROT gives an RDT a gain of 17.3\% in mAP when trained on derotated data and a CNN achieves mAP gains of 14.2\% when trained on underotated data but only a marginal gain of 2.5\% when trained on derotated data. We found that the confidence values for this specific case were not reliable (which directly effects mAP) because of confusion between fingertips (specifically index and ring) which further justified the creation of HandNet. \emph{For all experiments and datasets} the mP when using DeROT shows improvements of between 7.8\% and 21.1\% on underotated training data and between 23.5\% and 38.6\% for derotated training data. The simplistic Procrustes derotation negatively impacts fingertip detection relative to the baseline and we therefore chose not to build and train an RDT and CNN on Procrustes derotated versions of the two datasets. For our experiments a single RDT mostly outperforms a CNN. Although they are trained with different data and objectives it hints that there is no silver bullet to determining which machine learning approach is more appropriate.


\section{Conclusions and future work}\label{sec:conclusion}
We have shown that using derotation, specifically DeROT, significantly improves the localization ability of machine-learning based per-frame fingertip detectors by reducing the variance of the pose space. Furthermore we find that this procedure works despite the extremely high range of potential poses. We see this approach as an alternative to data augmentation and as a potentially useful additional step in pipelines dedicated to articulated object pose extraction such as hands. Although we have used no prior model or kinematic constraints to improve the detection results this is currently an active area that we are investigating. Also, in this work we have considered results only on depth images but it would be interesting to apply a similar pipeline to pure 2D color images. 
\\\\
\textbf{Acknowledgments}
This research was supported by European Community's FP7- ERC program, grant agreement no. 267414.

\bibliography{egbib}
\end{document}